\documentclass[accepted]{uai2026} % after acceptance, for a revised version; 
\usepackage[american]{babel}
\usepackage{natbib} % has a nice set of citation styles and commands
\usepackage{mathtools} % amsmath with fixes and additions
\usepackage{booktabs} % commands to create good-looking tables
\usepackage{tikz} % nice language for creating drawings and diagrams
\usepackage{hyperref}
\usepackage{microtype}
\usepackage{graphicx}
\usepackage{booktabs}
\usepackage{algorithm}
\usepackage{algorithmic}

\usepackage{amsmath}
\usepackage{amssymb}
\usepackage{mathtools}
\usepackage{amsthm}
\usepackage{bm}
\usepackage{subcaption}

\newtheorem{example}{Example} 
\newtheorem{theorem}{Theorem}
\newtheorem{lemma}[theorem]{Lemma} 
 
\newtheorem{remark}[theorem]{Remark}

\newtheorem{definition}[theorem]{Definition}

\newtheorem{assumption}[theorem]{Assumption}

\DeclareMathOperator*{\argmin}{arg\,min}

\newcommand{\E}{\mathbb{E}}
\newcommand{\R}{\mathbb{R}}
\newcommand{\Prob}{\mathbb{P}}

\newcommand{\N}{\mathbb{N}}

\newcommand{\price}{\mathrm{Price}}
\newcommand{\enr}{\mathrm{ENR}}
\newcommand{\tilmu}{\tilde{\mu}}
\newcommand{\muk}{\mu^{(k)}}
\newcommand{\mul}{\mu^{(l)}}
\newcommand{\Lap}{\mathrm{Lap}}
\newcommand{\supp}{\mathrm{supp}}

\title{Explainable Clustering of Mixture Models}

\author[1]{\href{mailto:<maximilian.fleissner@tum.de>?Subject=Your UAI 2026 paper}{Maximilian Fleissner}{}}
\author[1,2]{Maedeh Zarvandi}
\author[1,2]{Debarghya Ghoshdastidar}
\affil[1]{%
    Technical University of Munich
}
\affil[2]{
    Munich Center for Machine Learning
}
\begin{document}
\maketitle

\begin{abstract}%
    The explainable clustering problem was first posed by Moshkovitz et al. (ICML 2020) and studies how well an axis-aligned decision tree with $K$ leaves can approximate a given clustering. The performance of the tree is measured via the \textit{price of explainability}, defined as the ratio between the clustering cost of the tree (where every leaf is a cluster) and the optimal cost. Several recent works have given worst-case characterizations of the price of explainability for different cost functions. However, these guarantees are  data-agnostic and therefore notoriously pessimistic in practical clustering settings. In this paper, we study explainable clustering from the point of view of mixture models, which allows us to give the first data-dependent bounds on the price of explainability. First, we focus on $K$-medians clustering of mixture models with subexponential tails. We propose an algorithm that leverages information about the distribution of the data to find better cuts, and prove new upper and lower bounds. Second, we extend our algorithm and the theoretical guarantees it provides to kernel clustering, thereby refining the existing worst-case analysis. 
\end{abstract}

\section{Introduction}

Over the past decade, explainable machine learning has emerged as an active area of research, promising to cast light into black box algorithms. While several methods have been developed for the supervised setting, the explainability of unsupervised learning has attracted less attention, despite models deployed in practice often leveraging large amounts of unlabeled data. Recently however, several researchers have begun to address the problem of explainable $K$-means or $K$-medians clustering. In their ICML 2020 work, \citet{moshkovitz2020explainable} suggest to cluster the data using binary axis-aligned decision trees with $K$ leaves. At every node of the tree, a one sided axis-aligned threshold cut $x_i \le \theta$ partitions the data into two child nodes (where $x_i$ denotes the $i$-th coordinate of $x$). The clustering cost of this tree is then compared to the optimal $K$-medians or $K$-means solution. Of course, unless all clusters are incidentally already separable by axis-aligned cuts, the tree has a strictly higher cost. The quality of the approximation is measured by the \emph{price of explainability}, which is the ratio between the clustering cost induced by an optimal axis-aligned tree (where every leaf is a cluster) and the optimal clustering cost. \citet{moshkovitz2020explainable} prove an upper bound on the price of explainability that depends only on the number of clusters $K$, with $\mathcal{O}(K)$ guarantees for $K$-medians. This bound holds for \textit{any} dataset, regardless of how well it is clustered. 

Numerous recent works have given improved upper and lower bounds on the price of explainability for various cost functions (for an overview, see related works). Notably, \citet{makarychev2023random} show that for $K$-medians the price of explainability is $\Theta(\log K)$, and that the upper bound is attained by a randomized procedure known as Random Coordinate Cut. From a worst-case point of view, this settles the matter. However, all existing theory is entirely distribution-agnostic: It treats well-separated, low-noise mixture models and adversarial worst-case examples on equal footing.

This is overly pessimistic. In practice, it is unlikely that we would ever need to explain a clustering model that doesn't cluster the data well (instead, the data scientist would most likely be asked to fit a better model). Indeed, \citet{frost2020exkmc} empirically show that the price of explainability is actually close to $1$ for several real-world clustering datasets. Interestingly, already \citet{moshkovitz2020explainable} ponder in their discussion whether better guarantees are feasible for well-clustered data, such as Gaussian mixture models. On the other hand, as \citet{gamlath2021nearly} point out, the currently known examples which provide lower bounds on the price of explainability already consist of rather well-clustered instances, suggesting that care must be taken in the analysis.

\paragraph{Contributions.}

In this work, we study explainable clustering under assumptions on the data distribution, which allows us to derive tighter, more realistic guarantees. The first part of the paper tackles the classic explainable $K$-medians problem under mixture models with subexponential tails. We introduce an explainability-to-noise ratio $\enr(\nu)$ for mixtures $\nu$ and show that $\enr(\nu)$ influences how well a decision tree can approximate $\nu$. We prove upper and lower bounds on the price of explainability that depend only on the number of clusters $K$ and $\enr(\nu)$, and are almost tight in the latter. Our upper bounds are attained by the Mixture Model Decision Tree (MMDT) algorithm that we propose. MMDT takes specific information about the distribution of the mixture model into account and runs in data-independent time. The second part of this paper applies our new probabilistic analysis to kernel clustering. We derive data-dependent guarantees on the price of explainability that significantly improve over existing worst-case guarantees, thereby explaining the empirically observed phenomenon that the price of explainability is often quite low on real world clustering datasets.

\paragraph{Related Work.}

Starting with \citet{moshkovitz2020explainable}, a significant number of works have tackled the explainable clustering problem. Most works focus on a tighter analysis of the worst-case price of explainability for $K$-medians and $K$-means \citep{makarychev2021near, gamlath2021nearly, esfandiari2022almost, makarychev2023random, gupta2023price}. Some results also incorporate the role of the dimension $d$ into the analysis \citep{charikar2022near}, or focus on other cost functions \citep{laber2021price, laber2023nearly}. Other works on interpretable clustering loosen the requirement of exactly $K$ leaves. \citet{frost2020exkmc} empirically demonstrate that this improves the practical performance, and \citet{makarychev2022explainable} formally prove it. \citet{laber2023shallow} discuss explainable clustering with shallow decision trees. \citet{fleissner2024explaining} theoretically analyze the worst-case price of explainability for kernel clustering, focusing on Gaussian and Laplace kernel. Several other works also propose explainable clustering algorithms that can be used for partitions induced by kernel clustering, but do not bound the price of explainability \citep{bertsimas2021interpretable, ohl2024kernel, argov2026spex}.

\section{Problem Setup and Definitions}\label{sec:setup}

We begin by formalizing the model, discussing the problem statement, and introducing key definitions and notation.

\subsection{Data Model}

Let $K \ge 2$ be an integer. We are given a mixture model $\nu = \sum_{k=1}^K p^{(k)} \nu^{(k)}$, where each $\nu^{(k)}$ is a discrete or absolutely continuous probability distribution on $\R^d$, and $\sum_{k=1}^K p^{(k)} = 1$. We assume that the the mixing weights $p^{(k)}$ are safely bounded away from $1$, satisfying $p^{(k)} \le \frac{\alpha}{K}$ for some $\alpha \ge 1$. We denote by $\muk$ the mean of mixture component $\nu^{(k)}$, and assume that the probability density or probability mass function of each $\nu^{(k)}$ is symmetric around its mean, so that means and medians coincide for all $k$. We assume further that the mean absolute deviations of all mixture components are finite.
\begin{equation} \begin{aligned}
    \forall k, i: \E_{x \sim \nu^{(k)}}[|x_i - \mu_i^{(k)}|] = \sigma^{(k)}_i < \infty.
\end{aligned} \end{equation}
For all $i \in [d]$, we define the average mean absolute deviation as $\sigma_i = \sum_{k=1}^K p^{(k)} \sigma^{(k)}_i$. We use the superscript for enumerating the mixture components, and the subscript to refer to axes of the ambient space $\R^d$. The support of $\nu$ is denoted $\supp(\nu)$.

\begin{figure}[t]
\centering\includegraphics[width=\columnwidth]{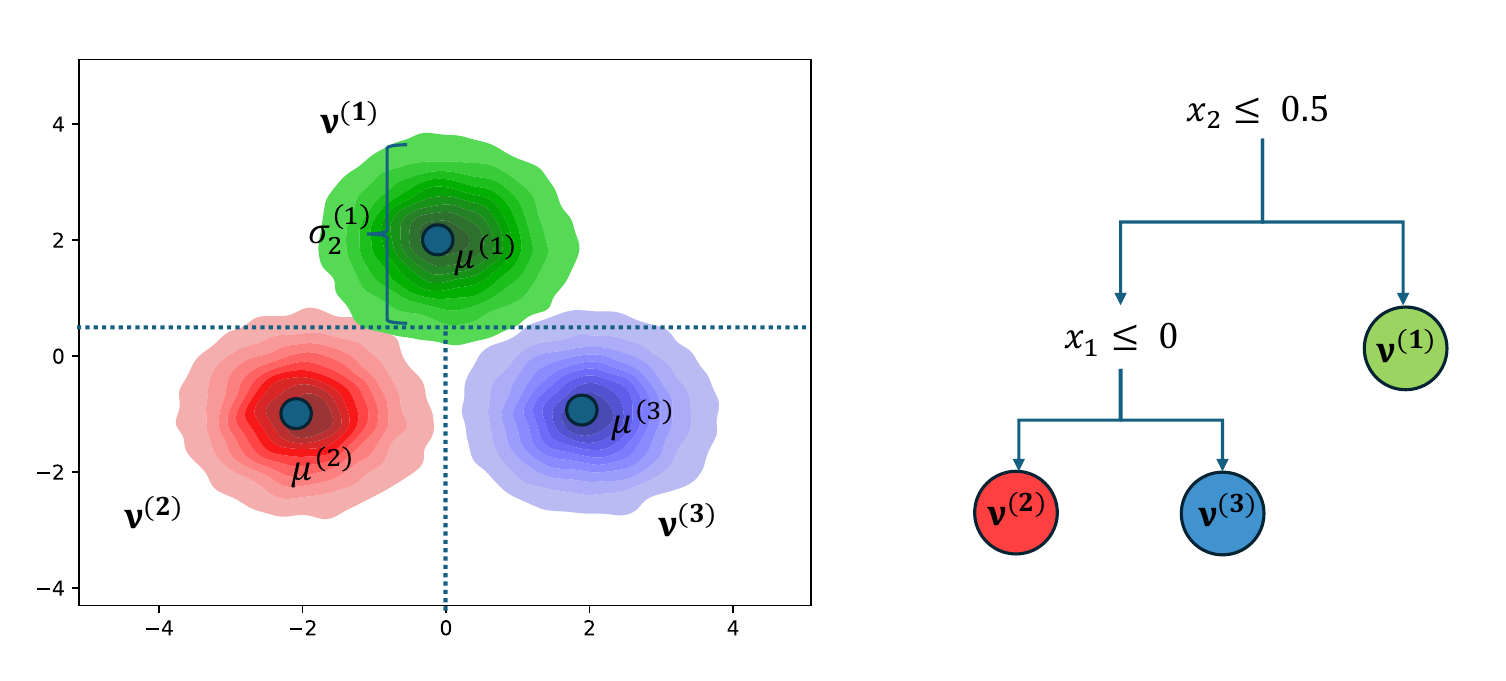}
    \caption{Illustration of our setting: Given three distributions $\nu^{(1)}, \nu^{(2)}, \nu^{(3)}$, we investigate how well a decision tree with axis-aligned binary cuts and three leaves can cluster these components. The objective is the ratio between the clustering cost of the tree, and the cost of the baseline partition into $\nu^{(1)}, \nu^{(2)}, \nu^{(3)}$.}\label{fig:illustration_task}
\end{figure}

\subsection{Problem Statement} Given a mixture model $\nu$ on $\R^d$, the goal is to explain each of its components in terms of individual features. In the explainable clustering problem, one chooses decision trees that sequentially partition the data into $K$ leaves, selecting at most $K$ features. Formally, given a mixture model $\nu$, our goal is therefore to construct an axis-aligned decision tree with $K$ leaves such that each leaf approximates exactly one of the $K$ mixture components $\nu^{(1)}, \dots, \nu^{(K)}$. At every internal node of the tree, a one-sided threshold cut $x_i \le \theta$ partitions the data into two child nodes. Here, we denote $x_i$ for the $i$-th coordinate of the point $x$. We ensure correspondence between leafs and mixture components by constraining the decision trees to satisfy that each leaf contains exactly one mean $\muk$. This is natural in a mixture model setting, but technically need not be enforced (there could exist trees with low clustering cost despite multiple means in the same leaf). However, all existing algorithms also demand every leaf to contain exactly one of the $K$ baseline cluster centers. Figure \ref{fig:illustration_task} illustrates our setting.

\subsection{Price of Explainability} 

We analyze the ratio between the cost of the tree and the cost of the underlying mixture model, defined as follows.

\begin{definition}[\textbf{Price of explainability for mixtures}]\label{def:price}
    Given a mixture model $\nu$ as specified above, and a decision tree $T$ that partitions $\R^d$ into $K$ leaves each containing one of the means $\muk$, we define 
    \begin{equation} \begin{aligned}\label{eq:price}
    \price(T,\nu) = \frac{\E_{x \sim \nu}\left[ \| x - \tilmu(x) \|_1 \right]}{\E_{x \sim \nu}\left[ \| x - \mu(x) \|_1 \right]} 
    \end{aligned} \end{equation}
    as the price of explainability. Here, $\mu(x) = \muk$ if $x$ is sampled from $\nu^{(k)}$, and $\tilmu(x)$ is the median of the leaf that $x$ ends up in. We denote further $\hat \mu(x)$ for the mean $\muk$ that ends up in the same leaf as $x$. This need not be $\tilmu(x)$.
\end{definition}

This is a direct adaption of the price of explainability to mixture models. However, the reference partition (i.e. the denominator) that we compare the decision tree against is the $K$-medians cost of assigning to each $x \sim \nu$ the median of the mixture component from which $x$ is drawn (i.e. one of the $\muk$). This is not necessarily the optimal $K$-medians partition associated with $\nu$. The reason is that the optimal centers of the $K$-medians cost 
\begin{equation} \begin{aligned}\label{eq:costopt}
    cost_{opt}(\nu) := \min_{c^{(1)}, \dots, c^{(K)}} \E_{x \sim \nu}\left[ \min_{k \in [K]} \|x-c^{(k)}\|_1\right]
\end{aligned} \end{equation}
are not given by the set $\{\muk\}_{k=1}^K$. As an example, think of $\nu$ being a mixture of two Gaussians in $\R$, with means $\mu^{(1)} = 1 = - \mu^{(2)}$ and unit variance. The minimizers of $cost_{opt}(\nu)$ are in general not $\pm 1$, because some $x \sim \nu$ will be sampled closer to the ``other'' mean, making it beneficial to push $c^{(1)}, c^{(2)}$ further apart. However, in a mixture model setting, we are typically interested in the components $\{\nu^{(k)}\}_{k=1}^K$ and not the optimal population clusters, and most works on mixture modeling aim to recover the true means and covariances, not the clustering \citep{dasgupta1999learning, ashtiani2020near}. In other words, our ground truth to compare an explainable clustering method against is the set of means $\{\muk\}_{k=1}^K$ of the mixture components, and we therefore stick to Definition \ref{def:price} in this paper. 

In this paper, we incorporate information on $\nu$ into the analysis of $\price(T,\nu)$.

\begin{definition}[\textbf{Explainability-to-noise ratio}]\label{def:enr}
For a mixture model $\nu$, we define its explainability-to-noise ratio as
\begin{equation} \begin{aligned}
\enr(\nu) := \min_{k \neq l} \max_{j \in [d]} \left(\frac{|\mu_j^{(k)} - \mu_j^{(l)}|}{\sigma_j} \right).
\end{aligned} \end{equation}
\end{definition}

A large explainability-to-noise ratio ensures that at any node of a decision tree that is not a leaf, we can find a pair of remaining means that can be separated along some axis $i \in [d]$ without increasing the probability $\Prob_{x \sim \nu}(\mu(x) \neq \hat \mu(x))$ too much. The explainability-to-noise ratio can be viewed as an axis-aligned version of the classic signal-to-noise ratio.

At this point, one may be tempted to think that if there exists a tree that makes few mistakes on a mixture model $\nu$, then the tree also automatically achieves a low price of explainability. However, this is not necessarily correct: The cost of the tree is not only influenced by the error rate of the tree, but also by the severity of the mistakes it makes. In some cases, both effects exactly ``cancel out''. For example, known $\Omega(\log K)$ lower bounds on the price of explainability for $K$-medians can be adapted such that the resulting tree almost perfectly recovers the mixture components $\nu^{(k)}$, in the sense that $\Prob(\hat \mu(x) = \mu(x)) = 1 -o(1)$. 

However, for mixture models with light tails, the error rate of the tree is indeed the deciding quantity, as it generally decays much faster than the cost of mistakes. This observation forms the starting point of our analysis and suggested algorithm.

\section{Proposed Algorithm}\label{sec:algo}

\begin{algorithm}[t]
\caption{Mixture Model Decision Tree (\textbf{MMDT})}
\label{alg:mmdt}
\small
\begin{algorithmic}[1]
\STATE \textbf{Input:} Components $\{\nu^{(k)}\}_{k=1}^K$ with means $\{\mu^{(k)}\}_{k=1}^K\subset \mathbb{R}^d$ and coordinate-wise mean absolute deviations $\{\sigma_i^{(k)}\}_{i,k}$.
\STATE \textbf{Output:} Binary tree with $K$ leaves (one per component).
\STATE Create root $t_0$ with $N(t_0)\gets [K]$, active set $\mathcal{A}\gets\{t_0\}$.
\WHILE{$\exists\, t\in\mathcal{A}$ with $|N(t)|>1$}
    \STATE Pick any such $t$, set $N\gets N(t)$.
    \STATE Choose $(i,\theta)$ by minimizing \eqref{eq:theta} over $i\in[d]$ and $\theta\in I_t$. If densities are unknown, by minimizing the bound in \eqref{eq:exponential} over $i,\theta$.
    \STATE Split index set into $N_L \gets \{k\in N:\mu^{(k)}_{i} \le \theta\}$ and $N_R \gets \{k\in N:\mu^{(k)}_{i} > \theta\}$.
    \STATE Store at $t$ the cut $(i,\theta)$, create children $t_L,t_R$ with $N(t_L)=N_L$, $N(t_R)=N_R$.
    \STATE Update $\mathcal{A}\gets(\mathcal{A}\setminus\{t\})\cup\{t_L,t_R\}$.
\ENDWHILE
\STATE Route $x$ at node $(i,\theta)$ to left if $x_i \le \theta$ and right otherwise.
\STATE If leaf $t$ has $N(t)=\{k\}$, assign leaf to component $k$.
\end{algorithmic}
\end{algorithm}

We propose the Mixture Model Decision Tree (MMDT) algorithm to obtain an explainable approximation for mixture models. MMDT is described in Algorithm \ref{alg:mmdt}. It iteratively partitions the means into $K$ leaves, each of which eventually contains exactly one mean, representing one mixture component. At every node $t$ with a set of remaining mixture components $N(t) \subset [K]$, this is achieved by identifying the axis $i \in [d]$ and binary cut $\theta \in \R$ such that the probability that any $x \sim \sum_{k \in N(t)} \bar{p}^{(k)} \nu^{(k)} =: \nu(N)$ is separated from its mean is minimal. Here, $\bar{p}^{(k)}$ simply rescales $p^{(k)}$ to ensure the weights of $\nu(N)$ still sum to one. Formally, 

\begin{equation}\label{eq:theta}
\begin{aligned}
&i,\theta 
= \argmin_{\substack{
    i \in [d], \theta \in I_t
}}
\Prob_{x \sim \nu(N)}\!\left(
    x \text{ separated from } \mu(x) \text{ by } \theta
\right) \\
&I_t = \left(\min_{k \in N(t)} \mu_i^{(k)}, \max_{k \in N(t)} \mu_i^{(k)}\right)
\end{aligned}
\end{equation}

If multiple possible thresholds exist, one is selected randomly. Typically, the exact probability density of each $\nu^{(k)}$ is not known (or difficult to minimize), so we choose $\theta$ by instead minimizing an upper bound on the probability in \eqref{eq:theta}. If nothing is known about the mixture components except its means and mean absolute deviations, we can use Markov's inequality: For any threshold $\theta$ chosen on the $i$-th axis, we can certainly bound the probability that $x \sim \nu^{(k)}$ gets separated from its mean by $\frac{|\mu_i^{(k)} -\theta|}{\sigma^{(k)}_i}$. We assume lighter, subexponential tails.

\begin{assumption}[\textbf{Subexponential tails}]\label{assumption_decay}
We assume there exist constants $c_1,c_2>0$ independent of $K$ such that for all $k \in [K], i \in [d], s>0$,
\begin{equation} \begin{aligned}
    \Prob_{x \sim \nu^{(k)}}( |x_i - \mu_i^{(k)}|>s ) \le c_1 \cdot \exp \left(-\frac{s c_2}{\sigma^{(k)}_i} \right).
\end{aligned} \end{equation}
\end{assumption}

Under Assumption \ref{assumption_decay}, for all $k \in [K]$, and any threshold cut $\theta$ along some axis $i \in [d]$, we can bound
\begin{equation}\label{eq:exponential}
\begin{aligned}
        &\Prob_{x \sim \nu^{(k)}}\left( x \text{ is separated from } \mu(x) \text{ by } \theta \right) \\ &\le c_1 \exp \left( -\frac{c_2|\theta - \mu_i^{(k)}|}{\sigma^{(k)}_i}\right).
\end{aligned}
\end{equation}
Plugging this bound into  \eqref{eq:theta}, $\theta$ can be found in data-independent time (i.e. without iterating over all datapoints).

\begin{remark} [\textbf{Connection to IMM}]
    Algorithm \ref{alg:mmdt} can be viewed as a ``population version" of IMM, the initial explainable clustering algorithm proposed by \citet{moshkovitz2020explainable}. IMM minimizes the number of points that go to the wrong cluster center at a given node. It admits data-independent $\mathcal{O}(K)$ bounds for $K$-medians, and its runtime scales linearly with the number of samples (unlike MMDT, which only relies on means and mean absolute deviations).
\end{remark}

\begin{remark}[\textbf{Improved runtime for high dimensions}]
    For high dimensions, finding the minimizer of \eqref{eq:exponential} over all $i \in [d]$ may be inefficient. For these cases, we suggest the following acceleration: At any node $t$ of the tree, and for all $i \in [d]$, compute
    \begin{equation} \begin{aligned}
        r_i(t) = \sigma_i^{-1} \left( \max_{l \in N(t)} \mul_i - \min_{k \in N(t)} \muk_i \right).
    \end{aligned} \end{equation}
    Then, for some small integer $d' \ll d$, only minimize \eqref{eq:exponential} over the axes $i_1, \dots, i_{d'}$ that achieve highest $r_i(t)$. Our theoretical guarantees are preserved even for $d'=1$.
\end{remark}

\begin{remark} [\textbf{Relaxing assumptions}]
    If the assumption of subexponential tails is considered too restrictive, one should replace Assumption \ref{assumption_decay} with a heavier-tailed decay rate. This in turn changes Equation \eqref{eq:exponential} and also the theoretical guarantees that can be attained. For the sake of clarity, we stick to Assumption \ref{assumption_decay} in the following section.
\end{remark}

\section{Theoretical Analysis of MMDT}\label{sec:theory}

In this section, we theoretically analyze Algorithm \ref{alg:mmdt} in terms of the price of explainability for $K$-medians. To simplify the analysis, we assume in this section that $\sigma_i = \sigma^{(k)}_i$ for all $i \in [d], k \in [K]$. 

\begin{theorem}[\textbf{Upper bounds on price}]\label{theo:price1} Consider a mixture model $\nu$ satisfying Assumption \ref{assumption_decay}. Denote $q = \enr(\nu)$. Then, Algorithm \ref{alg:mmdt} yields a decision tree $T$ such that
\begin{equation*} \begin{aligned}
    \price(T,\nu) \le 1 + \frac{c_1 \alpha \cdot q}{\exp \left(\frac{c_2 \cdot q }{2(K-1)} \right)-1}.
\end{aligned} \end{equation*}
\end{theorem}

The proof is in Appendix \ref{app:price1}.
Theorem \ref{theo:price1} quantifies the phenomenon that we alluded to in the introduction: \textit{Mixture models with light tails can usually be explained very well}, with price of explainability close to $1$ despite large $K$. We now provide a lower bound that is almost tight in $q$ for those trees that recover the individual mixture components sufficiently well.

\begin{theorem}[\textbf{Lower bounds on price}]\label{theo:price2} 
For any $K \ge 3$ and $q \ge (K-1) \log (1.5)$ there exists a mixture model $\nu$ with $K$ components, $\enr(\nu)=q$, equal mixing weights, and constants $c_1 = c_2 = 1$ such that the following holds: For any decision tree $T$ satisfying $\Prob_{x \sim \nu^{(k)}}( \hat \mu(x) = \mu^{(k)}) \ge 0.5$ for all $k$, and denoting $C_K = \frac{K!-K+1}{K!}$,
\begin{equation*} \begin{aligned}
    &\price(T,\nu) \ge \frac{C_K (2K-1)}{2K} + \frac{q}{ 96(K-1)  \cdot (e^{\frac{q}{K-1}}-1)}
\end{aligned} \end{equation*}
\end{theorem}

The proof is included in Appendix \ref{app:price2}. Since we need the constants $c_1, c_2$ to be independent of $K$, our construction differs from previous lower bounds on explainable $K$-medians.

\begin{remark}[\textbf{Comparison to Random Cuts}]\label{rem:compare}
    The Random Coordinate Cut algorithm analyzed by \citet{makarychev2023random} achieves $\mathcal{O}(\log K)$ worst case bounds for $K$-medians. In contrast, the upper bound for MMDT takes $\enr(\nu)$ into account. If $\enr(\nu) = \Theta(K)$, MMDT is loose by a factor of order $\Theta(\frac{K}{\log K})$ compared to Random Coordinate Cut. However, if the data is well-clustered and  $\enr(\nu) = \Omega( K \log K)$, the upper bound for MMDT is tighter. Being greedy at each node pays off if the data is easy to explain.
\end{remark}

\section{Extension to Kernel Clustering}\label{sec:kernels}

Kernel clustering is a nonparametric clustering technique based on the theory of reproducing kernel Hilbert spaces \citep{smola1998learning}. Intuitively, it relies on a suitable kernel function $\kappa: \R^d \times \R^d \rightarrow \R$ that measures the similarity between points $x,x' \in \R^d$. The potential nonlinearity of $\kappa$ allows discovering nonlinear cluster structures, which is impossible with ordinary $K$-means or $K$-medians. For kernel $K$-means with the Gaussian kernel $\kappa(x,x') = \exp(-\gamma \|x-x'\|^2)$ it has been shown that the price of explainability is essentially $\mathcal{O}(dK^2)$ in the worst-case. \textit{In this section, we extend our probabilistic analysis of explainable clustering to kernels, providing us with tighter, distribution-dependent guarantees}. The first step is to reformulate our problem in a nonparametric setting using kernel mean embeddings. We then introduce additional assumptions and definitions specific to this section.

\subsection{Background}

When a kernel $\kappa$ is positive semi-definite, there exists a Hilbert space $\mathcal{H}$ of functions and a feature map $\phi: \R^d \rightarrow \mathcal{H}$ such that $\kappa(x,x') = \langle \phi(x), \phi(x') \rangle$ for all $x,x' \in \R^d$. As before, let $\nu$ be a distribution on $\R^d$ with $\E_{x \sim \nu} \left[ \sqrt{\kappa(x,x)}\right] < \infty$. Then, the Bochner integral $\phi_\nu := \E_{x \sim \nu} \left[ \phi(x) \right] \in \mathcal{H}$ is referred to as the kernel mean embedding of the distribution $\nu$. For any $f \in \mathcal{H}$ it holds that $\E_{x \sim \nu} f(x) = \langle f, \phi_\nu \rangle$. Given two distributions $\nu, \nu'$, the maximum mean discrepancy (MMD) is defined as $d_{\kappa}( \nu, \nu') := \| \phi_\nu - \phi_{\nu'} \|_{\mathcal{H}}$. The MMD has numerous applications, such as two-sample testing \citep{JMLR:v13:gretton12a} and can be used to prove that kernel clustering recovers nonparametric mixture models \citep{vankadara2021recovery}. For more details on kernel mean embeddings, see \citet{muandet2017kernel}. 

\subsection{Assumptions}\label{subsec:kernel_assumptions}

The kernel setting is a bit different from the setup considered in the first part of this paper. Therefore, instead of the assumptions introduced in Sections \ref{sec:setup} and \ref{sec:algo}, we impose three conditions on the mixture $\nu$ and the kernel $\kappa$. Firstly, we assume that the kernel $\kappa$ is a bounded, distance-based product kernel
\begin{equation} \begin{aligned}\label{eq:productkernel}
    \kappa(x,x') = \prod_{i=1}^d \kappa_i(x,x') = \prod_{i=1}^d g_i(|x_i - x_i'|)
\end{aligned} \end{equation}
where each $g_i: [0,\infty) \rightarrow \R$ is a positive, monotone decreasing function satisfying $g_i(0)=1$. Examples include the Gaussian kernel with $g_i(t) = \exp(-t^2)$ for all $i$ and the Laplace kernel with $g_i(t) = \exp(-t)$ for all $i$. Secondly, we assume that the components $\nu^{(k)}$ of the mixture model $\nu$ all satisfy 
\begin{equation} \begin{aligned}\label{eq:normkme}
    \|\phi_{\nu^{(k)}} \|^2_\mathcal{H} = \E_{x,x' \sim_{i.i.d.} \nu^{(k)}} \left[ \kappa(x,x') \right] = \sigma^2
\end{aligned} \end{equation}
for some $\sigma^2 \in (0,1)$, and that $\nu^{(k)} \neq \nu^{(l)}$ for all $k \neq l$. Thirdly, we assume that for all $i \in [d], k \in [K]$ and $\bar x \in \supp(\nu)$ , the variance of $\kappa_i(\bar x, \cdot)$ is bounded by
\begin{equation} \begin{aligned}\label{eq:varkme}
    \mathbb{V}_{x \sim \nu^{(k)}} \left[ \kappa_i(\bar x, x) \right] \le \epsilon^2    
\end{aligned} \end{equation}
for some (small) $\epsilon > 0$. This is a concentration assumption on the coordinate-wise similarity between any fixed $\bar x$ and random $x \sim \nu^{(k)}$. Due to boundedness of the kernel, unbounded support of $\nu$ does not pose a problem, though the decay of the $g_i$ impacts how small we can hope $\epsilon$ to be.

Next, we extend Definition \ref{def:price} to kernels.

\begin{definition}[\textbf{Price of explainability in kernel clustering}]
    Consider a kernel function $\kappa$ satisfying \eqref{eq:productkernel}, and a mixture model $\nu$ satisyfing \eqref{eq:normkme} and \eqref{eq:varkme}. Then, we define the price of explainability of $\nu$ with respect to $\kappa$ as
    \begin{equation} \begin{aligned}
        \price_\kappa(T,\nu) &= \frac{\ E_{x \sim \nu} \left[ \| \phi(x) - \tilmu(x) \|^2_\mathcal{H} \right]}{E_{x \sim \nu} \left[ \| \phi(x) - \mu(x) \|^2_\mathcal{H} \right]}.
    \end{aligned} \end{equation}
    Here, $\tilmu(x)$ denotes the kernel mean embedding of the distribution $\nu$ conditioned on the event that it arrives in the $l$-th leaf of the decision tree. 
\end{definition}

\begin{remark}[\textbf{Upper bound on price for kernels}]
    We can certainly upper bound
    \begin{equation} \begin{aligned}
        \price_\kappa(T,\nu) &\le \frac{\ E_{x \sim \nu} \left[ \| \phi(x) - \hat \mu(x) \|^2_\mathcal{H} \right]}{E_{x \sim \nu} \left[ \| \phi(x) - \mu(x) \|^2_\mathcal{H} \right]}
    \end{aligned} \end{equation}
    where $\hat \mu(x) = \phi_{\nu^{(l)}}$ if $x$ ends up in a leaf that is assigned to the $l$-th mixture component $\nu^{(l)}$. This is true because the second moment functional $\mathcal{M}: \mathcal{H} \rightarrow \R$ defined via
    \begin{equation} \begin{aligned}
        \mathcal{M}(h) :=  \E \left[ \| \phi(x) - h\|^2_\mathcal{H} \right]
    \end{aligned} \end{equation}
    has its minimum at the mean $h = \E[\phi(x)]$.
\end{remark}

The explainability-to-noise ratio (Definition \ref{def:enr}) does not translate directly to the kernel setting, where clustering happens in $\mathcal{H}$ instead of $\R^d$. In the following, we instead give our guarantees on $\price_{\kappa}(T,\nu)$ in terms of the following axis-aligned quantity.

\begin{definition}[\textbf{Axis-aligned kernel distance}]\label{def:kappa_align} For a mixture model $\nu$ and a kernel $\kappa$ as introduced before, we define 
\begin{equation} \begin{aligned}
    \tau = \min_{k \neq l} \max_{i \in [d]} \E_{x,x' \sim \nu^{(k)},y \sim \nu^{(l)}}[\kappa_i(x,x')-\kappa_i(x,y)]
\end{aligned} \end{equation}
as the axis-aligned kernel distance of the mixture components.
\end{definition}

The axis-aligned kernel distance plays the role of the $\enr(\nu)$ in a nonparametric setting, except that is defined via inner products instead of distances (which is more natural when working with kernels). It can be upper bounded by the MMD between the mixture components, capturing the idea that well-clustered data is easier to explain even in kernel clustering. The following result is proved in Appendix \ref{app:lemma_mmd}. 

\begin{lemma}[\textbf{MMD implies bounds on $\tau$}]\label{lemma:lemma_mmd}
    Consider a mixture model $\nu$ such that for any two distinct $\nu^{(k)}, \nu^{(l)}$ it holds that $d_\kappa(\nu^{(k)}, \nu^{(l)}) \ge \gamma$. Then, $\tau \ge \frac{\gamma}{2d}$.
\end{lemma}

\subsection{Algorithm}

\begin{algorithm}[t]
\caption{Kernel Mixture Model Decision Tree (\textbf{Kernel MMDT})}
\label{alg:kmmdt}
\small
\begin{algorithmic}[1]
\STATE \textbf{Input:} Mixture components $\{\nu^{(k)}\}_{k=1}^K$ on $\R^d$, product kernel
$\kappa(x,x')=\prod_{i=1}^d \kappa_i(x,x')$ where $\kappa_i(x,x') = g_i(|x_i - x_i'|)$ and $g$ is monotone decreasing.
\STATE \textbf{Output:} Decision tree $T$ with $K$ leaves (one per component).

\STATE Root $t_0$ with $N(t_0)\gets [K]$, active set $\mathcal{A}\gets\{t_0\}$.
\WHILE{$\exists\, t\in\mathcal{A}$ with $|N(t)|>1$}
\STATE Pick any such $t$, set $N\gets N(t)$.
\STATE Find maximizers $(\bar x, i^*, k^*, l^*)$ of
\[
\xi(x,i,k,\ell)
:=\E_{x'\sim \nu^{(k)}}[\kappa_i(x,x')]
-\E_{y\sim \nu^{(\ell)}}[\kappa_i(x,y)].
\]
\STATE For each $m\in N$, compute $\zeta(m):=\E_{y\sim \nu^{(m)}}[\kappa_{i^*}(\bar x,y)]$.
\STATE If densities are known, choose $\rho \in (\min \zeta(m), \max \zeta(m))$ such as to minimize \eqref{eq:rho_choice}. If densities are unknown, select $\rho$ as midpoint of largest consecutive gap among sorted values $\{\zeta(m):m\in N\}$.
\STATE Split index set into $N_L:=\{m\in N:\zeta(m) \le \rho\}$ and $N_R:=\{m\in N:\zeta(m) > \rho\}$.
\STATE Define $\theta=g_{i^*}^{-1}(\rho)$.
\STATE Store at $t$ the cut $(i^*,\bar x_{i^*},\theta)$, create children $t_L,t_R$ with $N(t_L)=N_L$, $N(t_R)=N_R$.
\STATE Update $\mathcal{A}\gets(\mathcal{A}\setminus\{t\})\cup\{t_L,t_R\}$.
\ENDWHILE
\STATE At node $(i,\bar x_i,\theta)$ send $x$ right iff $|\bar x_i- x_i| < \theta$ (else left).
\STATE If $N(t)=\{k\}$, assign leaf to component $k$ and predict $\hat{\mu}(x)=\phi_{\nu^{(k)}}$.
\end{algorithmic}
\end{algorithm}

The main hurdle facing us in adapting the MMDT-algorithm to the kernel setting is that the kernel mean embeddings are no longer elements of $\R^d$. But for our decision tree to be interpretable, its axis-aligned cuts naturally need to be in the input space, not $\mathcal{H}$ (in general, the ``coordinates'' of $\mathcal{H}$ are functions, not coordinates of $\R^d$). Of course, one could in principle still operate on the means $\muk = \E_{x \sim \nu^{(k)}}[x] \in \R^d$. But these are in general not informative, because the clustering cost is measured in the Hilbert space. In fact, the input space means may even coincide despite the mixture components being well-clustered in $\mathcal{H}$.
 
\begin{example}
    For real numbers $r_1, \dots, r_K>0$ denote by $\nu^{(k)}$ the uniform distribution on the sphere of radius $r_k$ in $\R^d$. Let $\kappa$ be the Gaussian kernel. Then $\kappa(x,x') = \exp(-\|x-x'\|_2^2)  \le \exp(r_l-r_k)$ for all $x \in \supp(\nu^{(k)}), x' \in \supp(\nu^{(l)})$. However, $\E_{x \sim \nu^{(k)}}[x] = 0$ for all $k \in [K]$.
\end{example}

We resolve this by instead choosing \textit{prototype points} $\bar x$ at every node, and deciding based on these $\bar x$ whether a mixture component goes to the left or the right child node. The intuition is that, for at least one axis $i \in [d]$, $\kappa_i(\bar x,x')$ should be larger than some $\rho > 0$ for those $x'$ sampled from the same cluster as $\bar x$, and smaller otherwise. One thing changes compared to the previous section: Since
\begin{align}
    \kappa_i(\bar x ,x) \le \rho \iff |\bar x_i - x_i| \ge g_i^{-1}(\rho) =: \theta
\end{align}
the axis-aligned threshold cuts are no longer one sided cuts $x_i \le \theta$ but rather two-sided intervals $|x_i - \bar x_i| \le \theta$. This does not harm the interpretability of the resulting tree, as we are still only checking one axis at every node. This way, we can construct a decision tree in the input space, while at the same time relating its clustering cost to properties of the mixture components in $\mathcal{H}$.
The very same relaxation was done by \citet{fleissner2024explaining}.

As before, the tree $T$ is fitted sequentially. Suppose we have a set $N(t) \subset [K]$ of remaining mixture components at a node $t$. We choose an axis $i^* \in [d]$, a pair $k^*,l^* \in N(t)$, and a reference point $\bar x \in \supp( \nu^{(k^*)})$ such that
\begin{equation}\label{eq:xi_kernel} \begin{aligned}
    \xi(x,i,k,l) := \E_{x' \sim \nu^{(k)}, y \sim \nu^{(l)}}[ \kappa_i(x,x') - \kappa_i(x,y)]
\end{aligned} \end{equation}
is maximal. Note that certainly $\xi(\bar x, i^*,k^*,l^*) \ge \tau$ at each node if the axis-aligned kernel distance is $\ge \tau$. A threshold value $\rho$ is then decided as follows: For $m \in N(t)$, sort the terms
\begin{equation} \begin{aligned}
    \zeta(m) = \E_{y \sim \nu^{(m)}}[\kappa_{i^*}(\bar x,y)]
\end{aligned} \end{equation}
in non-decreasing order. If densities are known for each mixture component, then choose $\rho \in (\min \zeta(m), \max \zeta(m))$ such that
\begin{equation}\label{eq:rho_choice}
    \sum_{k \in N(t)} p^{(k)} \Prob_{x \sim \nu^{(k)}}\left( \kappa_i(\bar x,x), \zeta(k) \text{ lie on other sides of $\rho$}\right)
\end{equation}
is minimized. If densities are unknown, $\rho$ is chosen halfway between the two neighboring terms that are separated furthest. Now, for any $x \sim \nu$ that arrives at this node, we send it left if $\kappa_{i^*}(\bar x,x) < \rho$ and right otherwise. To keep track of which child node of $t$ receives the $m$-th mixture component, we simply evaluate on which side of $\rho$ the scalar $\zeta(m)$ is located. 

From there, the procedure continues at both child nodes. Eventually, each node only contains a single index $k \in [K]$. All points in this leaf are then assigned to the $k$-th mixture, that is $\hat \mu(x) = 
\phi_{\nu^{(k)}}$. We refer to this algorithm as Kernel MMDT, see Algorithm \ref{alg:kmmdt}.

\begin{remark}[\textbf{Choice of threshold in Kernel MMDT}]
    Similar as in MMDT, the threshold $\rho$ in Kernel MMDT is chosen with the aim of minimizing the probability that a point $x$ is separated from its corresponding mixture component. For large datasets, choosing the cut halfway in between two neighboring $\zeta(m)$ is a fast proxy.
\end{remark}

We can now give data-dependent guarantees on the price of explainability for kernels that depend on the axis-aligned kernel similarity $\tau$ and $\epsilon$, which measures the concentration within each cluster. The proof is in Appendix \ref{app:price_kernels}.

\begin{theorem}[\textbf{Price of explainability for kernels}]\label{theo:price_kernels}

Consider a mixture model $\nu$ as introduced above. Denote by $\tau>0$ its axis-aligned kernel distance. Then, it holds that
\begin{equation*} \begin{aligned}
        \price_\kappa(T,\nu) \le 1 + \frac{1 + \sigma^2}{1 - \sigma^2} \cdot \frac{(4 + \frac{2\pi^2}{3})\alpha \epsilon^2 (K-1)^2}{\tau^2}
\end{aligned} \end{equation*}
where $T$ is the tree fitted using the above algorithm (under known densities).
\end{theorem}

We conclude this section with two final remarks.

\begin{remark}[\textbf{Empirical Algorithm}]
For a finite dataset $\mathcal{X}= \{x^{(1)}, \dots, x^{(n)}\}$ of size $n$, the algorithm can be run on the empirical estimate $\hat \nu = \frac{1}{n} \sum_{i=1}^n \phi(x^{(i)})$ of $\nu$. Note that $\supp( \hat \nu)$ is finite, and every point $x^{(i)} \in \mathcal{X}$ is a potential prototype point to be selected by Kernel MMDT. If $n$ is large, we propose speeding the algorithm up by sampling a set of $S \in \N$ candidate prototypes $\mathcal{P} = \{ x^{(s,k)}\}_{s \in [S], k \in [K]}$.

\end{remark}

\begin{remark}[\textbf{Generalized threshold cuts}]
    As Lemma \ref{lemma:lemma_mmd} indicates, the axis-aligned kernel distance $\tau$ decreases with the input dimension $d$, suggesting that finding a ``good'' axis $i \in [d]$ is difficult in high dimensions, even when the mixture components are well-separated. For Gaussian and Laplace kernel, this can be resolved by allowing generalized threshold cuts of the form
    \begin{align}
        \kappa(\bar x, x) \le \rho \iff \| \bar x - x\|^p_p \ge -\log \rho.
    \end{align}
    where $p=1$ for Laplace and $p=2$ for Gaussian kernel. Of course, these cuts are no longer axis-aligned, but instead define $\ell_p$ balls around prototype points $\bar x$ in $\R^d$.
\end{remark}

\section{Experiments}\label{sec:experiments}

\begin{figure*}[!t]
\centering\includegraphics[width=0.95\linewidth]{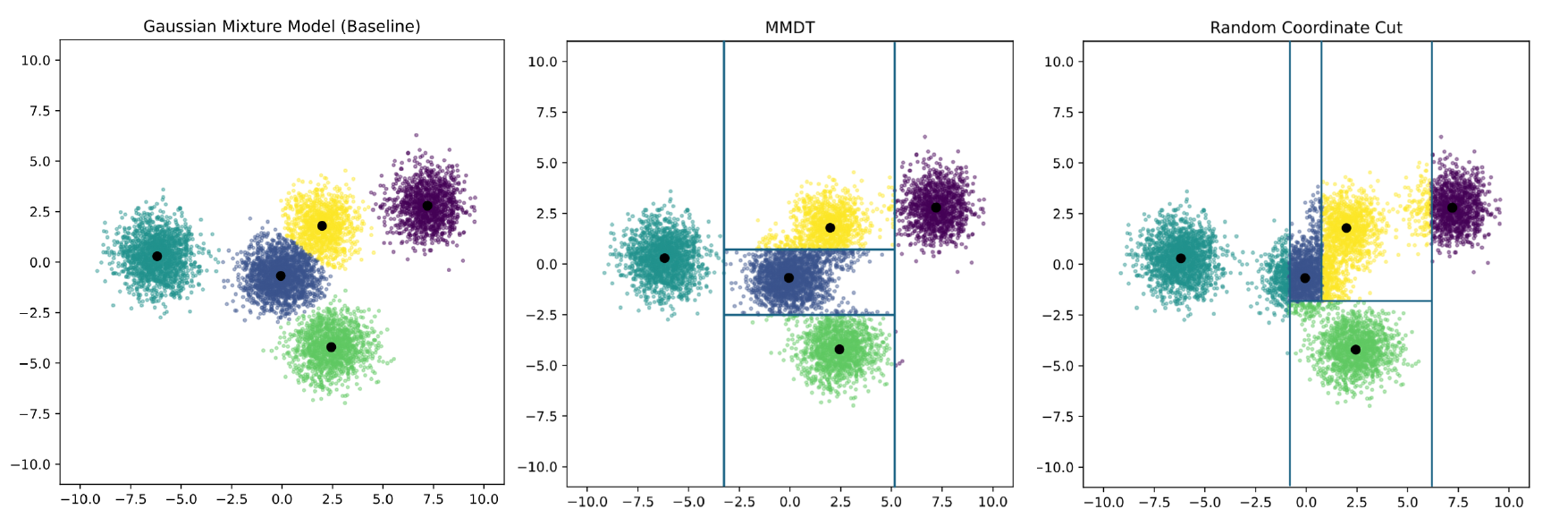}
\caption{On a Gaussian mixture model with $K=5$ components (left plot), MMDT exploits the concentration within each cluster to find an axis-aligned tree with low clustering cost (center plot). The distribution-agnostic Random Coordinate Cut randomly samples threshold cuts, leading to higher cost (right plot).}
\label{fig:gmm_compare}
\end{figure*}

\begin{table*}[!t]
\centering
\begin{tabular}{lrrrcccc}
\toprule
\textbf{Dataset} & \textbf{n} & \textbf{d} & \textbf{K} & \textbf{Price MMDT} & \textbf{Price CART} & \textbf{Price IMM} & \textbf{Average Price RCC} \\
\midrule
Gaussians I   & 7081  & 2  & 5  & 1.022 & 1.021 & 1.022 & 1.425 \\
Gaussians II  & 14970 & 10 & 10 & 1.000 & 1.010 & 1.000 & 1.967 \\
Wine          & 178   & 13 & 3  & 1.003 & 1.016 & 1.003 & 1.042 \\
Breast Cancer & 569   & 30 & 2  & 0.985 & 0.986 & 0.986 & 1.062 \\
\bottomrule
\end{tabular}
\caption{Price of explainability for four clustering datasets. On average over 500 runs, proposed MMDT outperforms RCC on all four datasets. The improvements are particularly noticeable for Gaussian mixture models. Note: The algorithm is not provided with the true clusters on all datasets, but instead estimates a mixture model.}
\label{table:experiments_kmedians}
\end{table*}

\begin{table*}[!t]
\centering
\begin{tabular}{lrrrccc}
\toprule
\textbf{Dataset} & \textbf{n} & \textbf{d} & \textbf{K} & \textbf{Price KMMDT} & \textbf{Price CART} & \textbf{Price SpeX} \\
\midrule
Iris         & 150  & 4    & 3 & 1.0742 & 1.0636 & 1.0636 \\
Digits       & 720  & 64   & 4 & 1.1240 & 1.0474 & 1.0263 \\
20newsgroups & 2756 & 1000 & 3 & 1.0045 & 1.0064 & 1.0040 \\
\bottomrule
\end{tabular}
\caption{We report the price of explainability for kernel clustering on three popular clustering datasets. We compare with CART and SpeX. The performance is comparably good for all three, indicating that the kernel clustering cost of the decision trees is close to the baseline partition.}
\label{table:experiments_kernel}
\end{table*}

To evaluate how robust our algorithms is, we compare it to IMM, CART, and Random Coordinate Cut on a number of synthetic and real world clustering datasets. First we look at two Gaussian mixture models (GMM) where the components have different covariance structures and non-uniform weights. The first GMM has $K=5$ clusters in $d=2$ dimensions, the second one has $K=10$ clusters in $d=10$ dimensions. Then, we run both algorithms on wine and the Wisconsin breast cancer dataset. $K$ is chosen based on the true number of classes in the dataset. 

The algorithms are \textbf{not} provided with the true mixture model but instead estimates weights, means and mean absolute deviations from the data using the GMM algorithm provided in scikit-learn \citep{scikit-learn}. The output of this algorithm is also used to compute the baseline $K$-medians cost. Note: Since the GMM algorithm may not find the optimal $K$-medians partition, it is possible that the price of explainability is $<1$ if the tree actually clusters slightly better than the estimated mixture model. We do not estimate the entire distribution, but use Equation \ref{eq:exponential} to compute the optimal threshold cut $(i,\theta)$ at each node. For Random Coordinate Cut, we report the average price of explainability over $500$ runs. \textit{MMDT outperforms the worst-case optimal RCC on all four datasets}. The difference is particularly pronounced when the assumption of a mixture model with light tails is satisfied. The full result is summarized in Table \ref{table:experiments_kmedians}. Figure \ref{fig:gmm_compare} shows an example with $K=5$ Gaussians in $\R^2$.

We also include empirical evaluations for Kernel MMDT on three datasets: Iris, MNIST digits 0-3, 20newsgroups.  For the latter, the features are constructed via tfidf. We use the Gaussian kernel with the bandwidth chosen proportional to the median Euclidean distance between points from the dataset, and compare with CART and SpeX \citep{argov2026spex}. As Table \ref{table:experiments_kernel} shows, Kernel MMDT performs comparably well, sometimes slightly weaker. Note that we do not claim to have derived the optimal explainable kernel clustering algorithm --- neither of the two competitors analyze price of explainability. Our contribution lies in showing that refined guarantees are possible by taking the data distribution into account.

\section{Discussion and Conclusion}\label{sec:discuss}

This paper provides the first analysis of the explainable clustering problem for mixture models. While a significant number of previous works have focused on worst-case bounds, we are the first to show that it is possible to obtain more realistic guarantees on the price of explainability by incorporating information on the data distribution.  It is noteworthy that our ideas extend to a kernel setting, and both the Gaussian as well as the Laplace kernel fall under the umbrella of our analysis. Given that both Gaussian and Laplace are characteristic kernels capable of distinguishing between arbitrary distributions \citep{sriperumbudur2008injective}, one might intuitively not expect interpretable approximations to exist. Our results provide a more nuanced look on this matter, demonstrating that axis-aligned cuts can work even for nonparametric mixture models provided the average similarity between samples from distinct mixture components is small, and dimensionality is not too high (see Lemma \ref{lemma:lemma_mmd}). 

There are a few open questions for future works to explore. For one, this paper assumes exact knowledge of the means and mean absolute deviations of the mixture model $\nu$, and studies \textit{approximation} guarantees for decision trees. This is the probabilistic equivalent of assuming knowledge of the true $K$-means or $K$-medians clusters, as is done in prior works. However, an immediate extension of our work would be to move towards \textit{generalization on new data} by incorporating finite sample complexity bounds on estimation of mixture models, e.g. when every component is Gaussian \citep{ashtiani2020near}. Since our proof mainly relies on the explainability-to-noise ratio $\enr(\nu)$ that depends only on means and mean absolute deviations, we do not believe that our results change for reasonably large sample sizes. Then again, it may not always be possible to exactly estimate $\nu$. This touches on the important question of identifiability in mixture models \citep{aragam2020identifiability}. 

% \begin{contributions} % will be removed in pdf for initial submission 
% 					  % (without ‘accepted’ option in \documentclass)
%                       % so you can already fill it to test with the
%                       % ‘accepted’ class option
%     Briefly list author contributions. 
%     This is a nice way of making clear who did what and to give proper credit.
%     This section is optional.

%     H.~Q.~Bovik conceived the idea and wrote the paper.
%     Coauthor One created the code.
%     Coauthor Two created the figures.
% \end{contributions}

\begin{acknowledgements}
This work is supported by the DAAD programme Konrad Zuse Schools of Excellence in Artificial Intelligence, sponsored by the Federal Ministry of Research, Technology and Space.
\end{acknowledgements}

% References
\bibliography{bib}

\newpage

\onecolumn

\title{Supplementary Material}
\maketitle

\appendix

\section{Proof of Theorem \ref{theo:price1}}\label{app:price1}

\begin{proof}
    Throughout this proof we assume $c_1 = c_2 = 1$ to avoid carrying all constants with us. Note that $\E_{x \sim \nu}\left[ \| x - \mu(x) \|_1 \right] = \sigma_1 + \dots + \sigma_d$. Moreover,
    \begin{align}
        \E_{x \sim \nu}\left[ \|x - \tilmu(x) \|_1 \right] \le \E_{x \sim \nu}\left[ \|x - \hat \mu(x) \|_1 \right]
    \end{align}
    because $\tilmu(x)$ is the median of the points that end up in the same leaf as $x$, and thus achieves the minimum possible $\| \cdot \|_1$ cost. Using the triangle inequality, we obtain 
    \begin{align}
        \E_{x \sim \nu}\left[ \| x - \hat \mu(x) \|_1 \right] \le \E_{x \sim \nu}\left[ \|x - \mu(x) \|_1 \right] + \E_{x \sim \nu}\left[ \| \mu(x) - \hat \mu(x) \|_1 \right].
    \end{align}
    Therefore, the price of explainability is given by
    \begin{align}
        \price(T,\nu) \le 1 + \frac{\E_{x \sim \nu}\left[ \| \mu(x) - \hat \mu(x) \|_1 \right]}{\sigma_1 + \dots + \sigma_d}.
    \end{align}
    The random variable $\| \mu(x) - \hat \mu(x) \|_1$ can be bounded uniformly from above.
    \begin{align}
        \| \mu(x) - \hat \mu(x) \|_1 \le \sum_{t \in T} \mathbf{1}(x \text{ is separated from } \mu(x) \text{ at node } t) \cdot \max_{k,l \in N(t)} \| \muk - \mul \|_1
    \end{align}
    where each $t \in T$ denotes a node of the tree $T$ and we bound $\| \mu(x) - \hat \mu(x) \|_1$ in terms of the worst-case distance between any pair from the set of remaining centers $N(t)$ at this very node. Taking expectations over $x \sim \nu$, we obtain
    \begin{align}
        \E_{x \sim \nu}\left[ \| \mu(x) - \hat \mu(x) \|_1 \right] \le \sum_{t \in T} \Prob(x \text{ is separated from } \mu(x) \text{ at node } t) \cdot \max_{k,l \in N(t)} \| \muk - \mul \|_1.
    \end{align}
    Let $t$ be an arbitrary node of the tree. For all $j \in [d]$, define
    $R_j(t) = \max_{k \in N(t)} \muk_j - \min_{l \in N(t)} \mul_j$ for the length of the interval that contains all remaining centers in $N(t)$, projected to the $j$-th axis. Note that there certainly exists $i \in [d]$ such that $R_{i}(t) \ge \sigma_{i} q$. This is true because $\enr(\nu) = q$.
    
    Recall that Algorithm \ref{alg:mmdt} chooses the threshold cut that \textit{minimizes} the probability of making a mistake at the node $t$. We can certainly upper bound this probability by analyzing only cuts along the particular axis $i \in [d]$.
    
    Observe that $x \sim \nu$ can only be separated from $\mu(x)$ at node $t$ along axis $i$ if it lies on the wrong side of the threshold cut $(i,\theta)$. This can only happen if $\left| x_i - \mu_i(x) \right| > \left| \theta - \mu_i(x) \right|$. Using Assumption \ref{assumption_decay}, the probability is therefore bounded as
    \begin{align}
        \Prob(x \text{ is separated from } \mu(x) \text{ at node } t) &= \sum_{k=1}^K p^{(k)} \cdot \Prob(x \text{ is separated from } \muk \text{ at node } t) \\
        &\le \frac{\alpha}{K} \sum_{k \in N(t)} \exp \left(-\frac{|\theta - \mu_i^{(k)}|}{\sigma_i} \right)
    \end{align}
    plugging in $p^{(k)} \le \frac{\alpha}{K}$.
    We assume w.l.o.g. that $N(t) = [K']$ for some $K' \le K$. We also assume that the projections are sorted in non-decreasing order, that is
    \begin{align}
        0 = \mu^{(1)}_i \le \mu^{(2)}_i \le \dots \le \mu^{(K')}_i = R_i(t).
    \end{align}
    Define the function
    \begin{align}
        f_i(\theta) = \sum_{k=1}^{K'} \exp \left(-\frac{|\theta - \mu_i^{(k)}|}{\sigma_i} \right).
    \end{align}
    This is an upper bound on the probability of separating any $x \sim \nu$ from its center at node $t$ through the threshold cut chosen by the algorithm.
    \\ \\
    \textbf{Claim 1:} Let $K'\ge 2$, $R_i(t)>0$, $\sigma_i>0$, and let $0=\mu_i^{(1)} \le \dots \le \mu_i^{(K')}=R_i(t)$.
    Define the function
    \begin{align}
        f_i(\theta)
        :=
        \sum_{k=1}^{K'}
        \exp\left(
        -\frac{|\theta-\mu_{j^*}^{(k)}|}{\sigma_{j^*}}
        \right),
    \end{align}
    and let $\theta^* \in \argmin_{\theta\in[0,R_i(t)]} f_i(\theta)$. Then, the following upper bound holds.
    \begin{equation}
    \begin{aligned}\label{eq:claim2}
    f_i(\theta^*)
    \le
    2 \sum_{k=1}^{\lceil K'/2 \rceil}
    \exp\left(
    -\frac{(2k-1)R_i(t)}
    {2(K'-1)\sigma_i}
    \right).
    \end{aligned}
    \end{equation}
    \\ \\
    \textbf{Proof of Claim 1:}
    The idea of the proof is to connect to Theorem 1.3 in \citep{farkas2025fenton}, which gives a very general result on minimax sums of translation-invariant functions. Set $m:=K'-1$, $R:=R_i(t)$, $\sigma:=\sigma_i$, and $\rho:=\frac{R}{\sigma}$. After the change of variables $s=\frac{\theta}{R}$ and with
    \begin{align}
    y_k:=\frac{\mu_i^{(k+1)}}{R}
    \end{align}
    for all $k=0,\dots,m$
    we obtain a normalized configuration
    \begin{align}
        0=y_0 \le \dots \le y_{m}=1
    \end{align}
    on the unit interval. We define
    \begin{align}
        &\Phi_y(s)
        :=
        \sum_{k=0}^{m} e^{-\rho |s-y_k|}, \\
        &f_i(\theta^*)
        =
        \inf_{s\in[0,1]} \Phi_y(s).
    \end{align}
    Then, taking the supremum over the points $y_k$,
    \begin{align}
        f_i(\theta^*)
        \le
        \sup_{0 \le y_1 \le \dots \le y_{m-1} \le 1} \;
        \inf_{s\in[0,1]} \; \Phi_y(s).
    \end{align}
    We now apply Theorem 1.3 of \citet{farkas2025fenton} to what is referred to as a non-singular kernel $K(u):=-e^{-\rho |u|}$ and a field function $J(s):=-e^{-\rho s}-e^{-\rho(1-s)}$. Note that $K(u)$ is piecewise concave and monotone on $(-1,0) \cup (0,1)$, and that $J(s)$ is finite everywhere (thus, for any $m$, it is a $m$-field function). Let
    \begin{align}
        F(y,s) := J(s) + \sum_{k=1}^{m-1} K(y_k-s).
    \end{align}
    Their result asserts that, in our notation,
    \begin{align}
       \inf_{0 \le y_1 \le \dots \le y_{m-1} \le 1} \; \sup_{s\in[0,1]} F(y,s) = \sup_{0 \le y_1 \le \dots \le y_{m-1} \le 1} \; \min_{0\le j\le m-1} \;
        \sup_{s\in[y_j,y_{j+1}]} F(y,s).
    \end{align}
    Note that $F(y,s) = - \Phi_y(s)$. By switching signs, this implies that
    \begin{align}
        \sup_{0 \le y_1 \le \dots \le y_{m-1} \le 1} \; \inf_{s\in[0,1]} \Phi_y(s)
         = 
        \inf_{0 \le y_1 \le \dots \le y_{m-1} \le 1} \; \max_{0\le j\le m-1} \;
        \inf_{s\in[y_j,y_{j+1}]}\Phi_y(s).
    \end{align}
    The left hand side is what we're after. The right hand side can be upper bounded by evaluating the respective $\max \inf$ objective for the equidistant grid
    $y_k =\frac{k}{m}$ where $k=0,\dots,m$. This grid places all points at a distance of $\frac{1}{m}$. For each interval $I_j(y):=\left[\frac{j}{m},\frac{j+1}{m}\right]$ where $j=0,\dots,m-1$,
    define the midpoint $s_j:=\frac{2j+1}{2m}$.
    Surely then, for all $j=0,\dots,m-1$, we have
    \begin{align}
        \inf_{s\in I_j(z)}\Phi_z(s)
        \le
        \Phi_z(s_j).
    \end{align}
    At such a midpoint, the two nearest grid points are at a distance of $\frac{1}{2m}$,
    the next two possible grid points are at distance of $\frac{3}{2m}$,
    and so on. Since there are only $m=K'-1$ grid points, we obtain
    \begin{align}
        \Phi_z(s_j)
        \le
        2\sum_{k=1}^{\lceil m/2\rceil}
        \exp\left(
        - \frac{\rho \cdot (2k-1)}{2m}
        \right).
    \end{align}
    This bound is independent of $j$. Consequently,
    \begin{align}
        \sup_y \inf_{s\in[0,1]}\Phi_y(s)
        \le
        2\sum_{k=1}^{\lceil m/2\rceil}
        \exp\left(
        - \frac{\rho \cdot (2k-1)}{2m}
        \right).
    \end{align} 
    Substituting back $m=K'-1$, and $\rho=\frac{R_i(t)}{\sigma_i}$ gives the claimed result. $\square$
    \\ \\
    Combining this result with the previous steps and using $K' \le K$, we see that at any node $t$, the probability of a mistake is bounded as follows.
    \begin{equation} \begin{aligned}\label{eq:prob_upper_bound_node}
        \Prob(x \text{ separated from } \mu(x) \text{ at } t)
        &\le \frac{2\alpha}{K} \sum_{k=1}^{\lceil K'/2 \rceil} \exp \left(-\frac{(2k-1)R_i(t)}{2(K'-1)\sigma_i} \right) \\
        &\le \frac{2 \alpha}{K} \sum_{k=1}^{\infty} \exp \left(-\frac{R_i(t)}{2(K'-1)\sigma_i} \right)^{(2k-1)} \\
        &\le \frac{2 \alpha}{K} \cdot \frac{1}{2} \cdot \sum_{k=1}^{\infty} \exp \left(-\frac{R_i(t)}{2(K'-1)\sigma_i} \right)^{k} \\
        &= \frac{\alpha}{K} \cdot \frac{\exp \left(-\frac{R_i(t)}{2(K'-1)\sigma_i} \right)}{1 - \exp \left(-\frac{R_i(t)}{2(K'-1)\sigma_i} \right)} \\
        &= \frac{\alpha}{K} \cdot \frac{1}{\exp \left(\frac{R_i(t)}{2(K-1)\sigma_i} \right)-1}
    \end{aligned} \end{equation}
    Returning to the price of explainability, we first note that
    \begin{equation} \begin{aligned}
        \max_{k,l \in N(t)} \| \muk - \mul \|_1 \le \sum_{j=1}^d R_j(t)
    \end{aligned} \end{equation}
    Therefore,
    \begin{equation} \begin{aligned}
        \frac{\E_{x \sim \nu}\left[ \| \mu(x) - \hat \mu(x) \|_1 \right]}{\E_{x \sim \nu} \left[ \| x - \mu(x) \|_1 \right] } &\le \sum_{t \in T} \frac{\sum_{j=1}^d R_j(t)}{\sum_{j=1}^d \sigma_j} \cdot \frac{\alpha}{K} \cdot \frac{1}{\exp \left(\frac{R_{i}(t)}{2(K-1)\sigma_{i}} \right)-1} \\
        &\le \sum_{t \in T} \frac{R_{i}(t)}{\sigma_{i}} \cdot \frac{\alpha}{K} \cdot \frac{1}{\exp \left(\frac{R_{i}(t)}{2(K-1)\sigma_{i}} \right)-1} \\
        &\le \sum_{t \in T} \frac{\alpha q}{K} \cdot \frac{1}{\exp \left(\frac{q}{2(K-1)} \right)-1} \\
        &\le \frac{\alpha q}{\exp \left(\frac{q}{2(K-1)} \right)-1}.
    \end{aligned} \end{equation}
In the third step, we used the fact that the function $x \mapsto \frac{x}{\exp(x)-1}$ is decreasing for all $x \ge 0$, and that $\frac{R_{i}(t)}{\sigma_{i}} \ge q$ by definition of the explainability-to-noise ratio. In the final step, we use that there are no more than $K-1$ nodes in the tree. 
\end{proof}

\section{Proof of Theorem \ref{theo:price2}}\label{app:price2}

\begin{proof}
    Fix any $K \in \N$ and $q \ge 1$. Define $\sigma = \frac{K-1}{q}>0$. We first construct means $\muk$. Let $d = K!$ and for all $j \in [d]$, let $\Pi_j$ be a permutation of the set $\{0,\dots,K-1\}$. Then, define $\muk_j = \Pi_j(k)$ for all $j \in [d], k \in [K]$. Thus, for all $l \neq k$, we have $\| \mul - \muk \|_1 \ge \frac{dK}{3}$. Let $\sigma = \frac{K-1}{q}$. For all $j \in [d], k \in [K]$, pick each $\nu^{(k)}$ to be a Laplace distribution (with independent marginals of scale $\sigma$) and mean $\muk$. This construction ensures that the mean absolute deviation is $\sigma$.
    Moreover, the distribution is subexponential with constants $c_1=1,c_2=1$. For all $j \in [d], k \in [K], s > 0$, standard results on the CDF for Laplace random variables give
    \begin{equation} \begin{aligned}
        \Prob_{x \sim \nu^{(k)}}(x_j \ge \mu_j^{(k)} + s) = \frac{1}{2} \exp \left( -\frac{s}{\sigma}\right).
    \end{aligned} \end{equation}
    We have uniform mixing weights, so $\nu = \frac{1}{K} \sum_{k=1}^K \nu^{(k)}$.
    Thus, $\enr(\nu) = \frac{K-1}{\sigma} = q$ and $\E[\|x-\mu(x)\|_1] = \sum_{j=1}^d \E[|x_j-\mu_j(x)|] = d \sigma$. Since the distribution is symmetric, means and medians coincide, so this is also the baseline $K$-medians cost. Now consider as tree $T$ that puts every $\muk$ in its own leaf. At its root, the probability of putting a sample $x \sim \nu$ to a leaf that dos not contain $\mu(x)$ is at least
    \begin{equation} \begin{aligned}
        \Prob(x\text{ separated from } \mu(x)) &\ge \frac{1}{2K} \sum_{k=1}^{K-1} \exp \left( -\frac{k}{\sigma}\right) \\
        &= \frac{1}{2K} \frac{e^{-\frac{1}{\sigma}} - e^{-\frac{K}{\sigma}}}{1 - e^{-\frac{1}{\sigma}}} \\
        &\ge \frac{1}{4K} \frac{1}{e^{\frac{1}{\sigma}}-1}.
    \end{aligned} \end{equation}
    The last line uses $q \ge (K-1) \log 1.5 \ge \log 2$. 
    Denote by $\tilmu(x)$ the median of the leaf that a sample $x$ ends up. Denote by $\mathcal{I}(T) \subset [d]$ the coordinates used in a tree $T$. To lower bound the price of explainability, we will rely on the lower bound
    \begin{align}
        \E_{x \sim \nu}[\|x-\tilmu(x)\|_1 \ge \sum_{j \notin \mathcal{I}(T)} \E_{x \sim \nu}[|x_j-\tilmu_j(x)|].
    \end{align}
    Note: For $j \notin \mathcal{I}(T)$, the distribution of $x_j$ is independent of any events that depend on the tree (such as whether the point was split in other axes). Therefore, we can later condition on these events without changing the expectation.
    
    For all $l \neq k$, the distance between the means restricted to the coordinates $j \notin \mathcal{I}(T)$ is still of order $dK$. This is true because the worst that can happen for any pair $l \neq k$ is that the coordinates in $\mathcal{I}(T)$ (of which there are no more than $K-1$ since the depth of the tree is at most $K$) remove exactly $K-1$ of those $2(K-2)!$ coordinates where the distance between $\muk$ and $\mul$ is $K-1$. Thus,
    \begin{equation} \begin{aligned}
        \sum_{j \notin \mathcal{I}(T)} |\muk_j - \mul_j| &\ge \frac{dK}{3} - (K-1)^2 \ge dK \cdot \left( \frac{1}{3} - \frac{K-1}{K^2} \right) \ge \frac{dK}{12}.
    \end{aligned} \end{equation}
    This already shows a weaker result, namely that
    \begin{equation} \begin{aligned}
        \frac{\E_{x \sim \nu}[ \| x - \hat \mu(x) \|_1]}{\E_{x \sim \nu}[ \| x -  \mu(x) \|_1]} &\ge \frac{1}{d \sigma} \cdot \Big( \sum_{j \notin \mathcal{I}(T)} P(\hat \mu(x) = \mu(x)) \cdot \E[|x_j - \mu(x)_j|] + P(\hat \mu(x) \neq \mu(x)) \cdot \frac{dK}{12} \Big) \\
        &\ge \frac{d-K+1}{d} + \frac{K}{12 \sigma} \cdot \frac{1}{4K} \frac{1}{e^{\frac{1}{\sigma}}-1} \\
        &= \underbrace{\frac{d-K+1}{d}}_{  \approx 1 \text{ for large $K$}} + \frac{q}{48 (K-1)} \cdot  \frac{1}{e^{\frac{q}{K-1}}-1}.  
    \end{aligned}
    \end{equation}
    To proceed further, we need to account for the fact that $\hat \mu(x) \neq \tilmu(x)$. We start with a lower bound on the median of a sum of Laplace random variables with ``distant'' means.
    \\ \\
\textbf{Claim 2:} Let $K \in \N, p \in \N$ and $ \alpha_1, \dots, \alpha_K \in \R^p$, and $\sigma>0$. For all $k \in [K]$ let $L_k \sim \Lap(\alpha_k,\sigma)$ denote the $p$-dimensional Laplace distribution with mean $\alpha_k \in \R^p$ and independent marginals of scale $\sigma$. Suppose that there exists $C>0$ such that $\| \alpha_k - \alpha_l \|_1 > C$ for all $l \neq k$. Let $p_1 \ge p_2 \ge \dots \ge p_K \ge 0$ denote mixing weights that sum to one, and let $m \in \R^p$ be the median of the mixture. Then,
\begin{align}
    \sum_{k=1}^K p_k \E[ \|L_k - m \|_1]
    \ge p_1 \cdot p\sigma + (1-p_1)\left( 0.5C + p \sigma e^{-\frac{C}{2p\sigma}}\right).
\end{align}

\textbf{Proof of Claim 2:}
For $k \in [K]$, define $r_k := \|\alpha_k-m\|_1$.
Since the coordinates of $L_k$ are independent Laplace random variables with scale $\sigma$, the one-dimensional identity $\E_{X \sim \Lap(\mu,\sigma)}[|X-a|] = | \mu-a| + \sigma \exp\left(-\frac{| \mu-a|}{\sigma}\right)$ yields 
\begin{equation}
\begin{aligned}
    \E[\|L_k-m\|_1]
    &= \| \alpha_k - m\|_1 + \sigma \sum_{j=1}^p \exp \left(-\frac{|e_j^\top \alpha_{k}-m_j|}{\sigma} \right) \\
    &\ge \| \alpha_k - m\|_1 + p \sigma \cdot \exp \left( -\frac{\| \alpha_k - m\|_1}{p\sigma} \right).
\end{aligned}
\end{equation}
In the second step we used the AM-GM inequality, i.e. $\sum_{i=1}^n u_i \ge n \prod_{i=1}^n u_i^{1/n}$ for any $u_1, \dots, u_n \ge 0$.
Therefore,
\begin{equation}
\begin{aligned}
    \E[\|L_k-m\|_1]
    \ge r_k + p\sigma \exp\left(-\frac{r_k}{p\sigma}\right) =  \psi(r_k),
\end{aligned}
\end{equation}
where we define the function $\psi(t) := t + p \sigma \exp (-\frac{t}{p\sigma})$ for all $t \ge 0$. Observe that $\psi(0)=p \sigma$ and $\psi'(t) = 1 - \exp(-\frac{t}{p \sigma}) \ge 0$. Thus, $\psi$ is non-decreasing. Let $i \in [K]$ be the index of the mean $\alpha_i$ that is closest to the median. We write
\begin{align}
    r_i = \min_{k \in [K]} r_k =: c.
\end{align}
We distinguish two cases.
\begin{enumerate}
    \item If $c \ge 0.5C$, then $r_k \ge 0.5C$ for all $k$, and thus
\begin{equation}
\begin{aligned}
    \sum_{k=1}^K p_k \psi(r_k)
    &\ge \psi\left(0.5C\right) = 0.5C + p \sigma \cdot \exp \left(-\frac{0.5 C}{p\sigma} \right) \ge p_1 \cdot p\sigma + (1-p_1) \left( 0.5C + p \sigma e^{-\frac{C}{2p \sigma}}\right).
\end{aligned}
\end{equation}
\item If $c < 0.5C$, then for all \textbf{other} $k \neq i$, the triangle inequality gives
\begin{equation}
\begin{aligned}
    r_k
    = \|\alpha_k-m\|_1
    &\ge \|\alpha_k-\alpha_i\|_1 - \|\alpha_i-m\|_1 > C-c \ge 0.5C.
\end{aligned}
\end{equation}
Therefore, using that $\psi$ is non-decreasing and that $p_1$ is the largest weight,
\begin{equation}
\begin{aligned}
    \sum_{k=1}^K p_k \psi(r_k)
    &\ge p_i \psi(r_i) + \sum_{k \neq i} p_k \psi(r_k) \\
    &\ge p_i \psi(0) + \sum_{k \neq i} p_k \psi\left(0.5C\right) \\
    &\ge p_1 \psi(0) + \sum_{k \neq 1} p_k \psi\left(0.5C\right) \\
    &= p_1 p \sigma + (1-p_1) \cdot \left( 0.5C + p \sigma \exp \left( - \frac{0.5C}{p \sigma} \right)\right).
\end{aligned}
\end{equation}
\end{enumerate}
This proves the claim. $\square$ \\ \\
We are ready to complete the proof. Let $p_{k \to l} = \mathbb{P}(\mu(x) = \muk, \hat \mu(x) = \mul)$ denote the probability mass of component $k$ assigned to leaf $l$. Summing the cost over all leaves and using the law of total expectation,
\begin{equation} \begin{aligned}
\E_{x \sim \nu}[\|x - \tilde{\mu}(x)\|_1]
&= \sum_{l,k=1}^K \Prob_{x \sim \nu}(\mu(x) = \muk, \hat \mu(x)=\mul) \cdot \mathbb{E}_{x \sim \nu} \Big[\|x - \tilmu^{(l)}\|_1 \Big| \mu(x) = \muk, \hat \mu(x)=\mul \Big] \\
&\ge \sum_{j \notin \mathcal{I}(T)} \sum_{l,k=1}^K \Prob_{x \sim \nu}(\mu(x) = \muk, \hat \mu(x)=\mul) \cdot \mathbb{E}_{x \sim \nu^{(k)}}\Big[|x_j - \tilmu_j^{(l)}| \Big| \hat \mu(x)=\mul \Big] \\
&= \sum_{l,k=1}^K p_{k \to l} \sum_{j \notin \mathcal{I}(T)} \mathbb{E}_{x \sim \nu^{(k)}}[|x_j - \tilmu_j^{(l)}|]
\end{aligned} \end{equation}
We used the fact that for $x \sim \nu^{(k)}$, the random variable $x_j$ is independent of the event $\hat \mu(x) = \mul$ for $j \notin \mathcal{I}(T)$. As stated before, for all $k \neq k'$, we have $\sum_{j \notin \mathcal{I}(T)} |\muk_j - \mu^{(k')}_j| \ge \frac{dK}{12} =: C$. Now, we may invoke Claim 2 for all leaves $l \in [K]$. Define $\bar p_l := \sum_{k=1}^K p_{k \to l}$ for the probability mass of the $l$-th leaf. Recall that we restrict ourselves to trees $T$ satisfying
\begin{equation}\label{eq:lower_bound_error_strengthened}
\begin{aligned}
    \forall k \in [K]: \Prob_{x \sim \nu^{(k)}}(\hat \mu(x) = \mu(x)) \ge \frac{1}{2}.
\end{aligned}
\end{equation}
Due to equal mixing weights, this implies $p_{l \to l} \ge \frac{1}{2K}$ for all $l$. Furthermore, for all $l \neq k$, $p_{k \to l} \le \frac{1}{2K}$ and hence $\max_{k} p_{k \to l} = p_{l \to l}$. Define $p =d- |\mathcal{I}(T)|$. Then,
\begin{equation}
\begin{aligned}
 \sum_{l,k=1}^K p_{k \to l} \sum_{j \notin \mathcal{I}(T)} \mathbb{E}_{x \sim \nu^{(k)}}[|x_j - \tilmu_j^{(l)}|]
 &= \sum_{l=1}^K \bar p_l \cdot \sum_{k=1}^K \frac{p_{k \to l}}{\bar p_l}  \sum_{j \notin \mathcal{I}(T)} \mathbb{E}_{x \sim \nu^{(k)}}[|x_j - \tilmu_j^{(l)}|] \\
 &\ge \sum_{l=1}^K \bar p_l \cdot \left( \frac{p_{l \to l}}{\bar p_l} \cdot p \sigma + \left(1-\frac{p_{l \to l}}{\bar p_l} \right) \cdot \left(0.5C + p \sigma e^{-\frac{0.5C}{p \sigma}} \right)\right) \\
 &= \sum_{l=1}^K p_{l \to l} \cdot p \sigma + (\bar p_l - p_{l \to l}) \cdot \left(0.5C + p \sigma e^{-\frac{0.5C}{p \sigma}} \right) \\
 &= p \sigma + \sum_{l=1}^K (\bar p_l - p_{l \to l}) \cdot \left( 0.5C + p \sigma e^{-\frac{0.5C}{p \sigma}} - p \sigma \right)
\end{aligned}
\end{equation}
where we used $\sum_{l=1}^K \bar p_l = 1$ in the last line. Furthermore,
\begin{equation}
\begin{aligned}
    \sum_{l=1}^K \bar p_l - p_{l \to l}
    = \sum_{l=1}^K \sum_{k \neq l} p_{k \to l} = \Prob(\hat \mu(x) \neq \mu(x))
    \ge \frac{1}{4K} \frac{1}{e^{\frac{1}{\sigma}}-1}.
\end{aligned}
\end{equation}
Plugging this into the bound,
\begin{equation}
\begin{aligned}
\E_{x \sim \nu}[\|x - \tilde{\mu}(x)\|_1]
&\ge p \sigma + \frac{1}{4K} \cdot \frac{0.5C + p \sigma e^{-\frac{0.5C}{p \sigma}} - p \sigma}{e^{\frac{1}{\sigma}}-1}.
\end{aligned}
\end{equation}
Recall $C=\frac{dK}{12}$ and $q = \frac{K-1}{\sigma}$, so $\frac{1}{\sigma} = \frac{q}{K-1}$ and $\frac{0.5 C}{4d \sigma K} = \frac{q}{96(K-1)}$. Let $C_K := \frac{K!-K+1}{K!}$ as defined in the theorem. Note: $\frac{p}{d} \ge C_K$ since $p \ge d-K+1$. Dividing the cost of the tree by the baseline cost $d \sigma$, some algebra gives
\begin{equation}
\begin{aligned}
    \price(T,\nu)
    &\ge 
    C_K
    +
    \frac{1}{e^{\frac{q}{K-1}}-1}
    \left(
        \frac{q}{96(K-1)}
        +
        \frac{p}{4Kd}
        \left(
            e^{-\frac{C}{2p\sigma}} - 1
        \right)
    \right) \\
    &\ge 
    C_K
    +
    \frac{1}{e^{\frac{q}{K-1}}-1}
    \left(
        \frac{q}{96(K-1)}
        -
        \frac{p}{4Kd}
    \right) \\
    &\ge 
    C_K
    -
    \frac{p}{2Kd}
    +
    \frac{q}{96(K-1)\left(e^{\frac{q}{K-1}}-1\right)} \\
    &\ge 
    \frac{C_K(2K-1)}{2K}
    +
    \frac{q}{96(K-1)\left(e^{\frac{q}{K-1}}-1\right)} .
\end{aligned}
\end{equation}
where $p \le d$ and $q \ge (K-1) \log 1.5$ are used. This completes the proof.
\end{proof}

\section{Proof of Lemma \ref{lemma:lemma_mmd}}\label{app:lemma_mmd}

\begin{proof}
    Recall that for all $k \in [K]$, we have
    \begin{equation} \begin{aligned}
        \E_{x,x' \sim_{i.i.d.} \nu^{(k)}} \left[ \kappa(x,x') \right] = \| \phi_{\nu^{(k)}} \|^2_\mathcal{H} = \sigma^2.
    \end{aligned} \end{equation}
    Now take any pair of distinct mixture components $\nu^{(k)}, \nu^{(l)}$. First of all, we have
    \begin{align}
        \gamma &\le d_\kappa(\nu^{(k)}, \nu^{(l)}) \\
        &= \E_{x,x' \sim_{i.i.d.} \nu^{(k)}} \left[ \kappa(x,x') \right] + \E_{y,y' \sim \nu^{(l)}} \left[ \kappa(y,y') \right]  - 2\E_{x \sim \nu^{(k)}, y \sim \nu^{(l)}} \left[ \kappa(x,y) \right] \\
        &= 2\E_{x,x' \sim_{i.i.d.} \nu^{(k)}} \left[ \kappa(x,x') \right] - 2 \E_{x \sim \nu^{(k)}, y \sim \nu^{(l)}} \left[ \kappa(x,y) \right].
    \end{align}
    Therefore, we can write
    \begin{equation} \begin{aligned}\label{eq:kappa_i_difference}
        \gamma/2
        &\le \E_{x,x' \sim_{i.i.d.} \nu^{(k)}} \left[ \kappa(x,x') \right] - \E_{x \sim \nu^{(k)}, y \sim \nu^{(l)}} \left[ \kappa(x,y) \right] \\
        &= \E_{x,x' \sim_{i.i.d.} \nu^{(k)}, y \sim \nu^{(l)}} \left[ \prod_{i=1}^d g_i(|x_i - x_i'|) - \prod_{i=1}^d g_i(|x_i - y_i|) \right].
    \end{aligned} \end{equation}
    Next, we use the fact that for any set of positive real numbers $a_i,b_i \in (0,1]$ we can write
    \begin{equation} \begin{aligned}
    \prod_{i=1}^d a_i - \prod_{i=1}^d b_i &= (a_1 - b_1) \prod_{i=2}^d a_i +
    b_1 (a_2 - b_2) \prod_{i=3}^d a_i +
    b_1 b_2 (a_3 - b_3) \prod_{i=4}^d a_i + \dots \\
    &= \sum_{i=1}^d (a_i - b_i) \underbrace{\prod_{j < i} b_j}_{\in (0,1]} \cdot \underbrace{\prod_{j > i} a_j}_{\in (0,1]} \\
    &\le \sum_{i: a_i \ge b_i} (a_i - b_i).
    \end{aligned} \end{equation}
    Therefore, there must exist $i \in [d]$ with
    \begin{equation} \begin{aligned}
        a_i - b_i \ge \frac{\prod_{i=1}^d a_i - \prod_{i=1}^d b_i}{d}.
    \end{aligned} \end{equation}
    It follows from Equation \eqref{eq:kappa_i_difference} that there certainly exists $i \in [d]$ with
    \begin{equation} \begin{aligned}
        \E_{x,x' \sim_{i.i.d.} \nu^{(k)}, y \sim \nu^{(l)}} \left[  g_i(|x_i - x_i'|) - g_i(|x_i - y_i|) \right] \ge \frac{\gamma}{2d}.
    \end{aligned} \end{equation}
    Since the pair $k \neq l$ was arbitrary, this proves the statement.
    
\end{proof}

\section{Proof of Theorem \ref{theo:price_kernels}}\label{app:price_kernels}

\begin{proof}
    We begin by noting that for any $x \sim \nu$, if $\hat \mu(x) \neq \mu(x)$, then
    \begin{equation} \begin{aligned}
        \|\phi(x) - \hat \mu(x) \|_\mathcal{H}^2 = \| \phi(x)\|^2_\mathcal{H} + \|\hat \mu(x) \|_\mathcal{H}^2 - 2 \langle \phi(x), \hat \mu(x) \rangle \le 1 + \sigma^2.
    \end{aligned} \end{equation}
    Here, we plugged in $\| \phi(x)\|^2_\mathcal{H} = \kappa(x,x) = 1$ and then used the fact that $\hat \mu(x)$ is a kernel mean embedding to get
    \begin{equation} \begin{aligned}
        \langle \phi(x), \hat \mu(x) \rangle = \E_{y \sim \nu^{\hat \mu(x)}} [ \kappa(x,y) ] \ge 0.
    \end{aligned} \end{equation}
    Thus, 
    \begin{equation} \begin{aligned}
       \| \phi(x) - \hat \mu(x) \|^2_\mathcal{H} \le  \| \phi(x) - \mu(x) \|^2_\mathcal{H} + \bm 1( \hat \mu(x) \neq \mu(x)) \cdot (1 + \sigma^2).
    \end{aligned} \end{equation}
    Therefore, 
    \begin{equation} \begin{aligned}
        \E_{x \sim \nu} \left[ \| \phi(x) - \hat \mu(x) \|^2_\mathcal{H} \right] \le \E_{x \sim \nu} \left[ \| \phi(x) - \mu(x) \|^2_\mathcal{H} \right] + \Prob( \hat \mu(x) \neq \mu(x)) \cdot (1+\sigma^2).
    \end{aligned} \end{equation}
    This implies that
    \begin{equation} \begin{aligned}
        \frac{ \E_{x \sim \nu} \left[ \| \phi(x) - \hat \mu(x) \|^2_\mathcal{H} \right]}{\E_{x \sim \nu} \left[ \| \phi(x) - \mu(x) \|^2_\mathcal{H} \right]}
        &\le \frac{\E_{x \sim \nu}\left[ \| \phi(x) - \mu(x) \|_\mathcal{H}^2 \right] + \Prob( \hat \mu(x) \neq \mu(x)) \cdot (1+\sigma^2)}{\E_{x \sim \nu} \left[ \| \phi(x) - \mu(x) \|^2_\mathcal{H} \right]} \\
        &= 1 + \frac{1+\sigma^2}{1-\sigma^2} \cdot \Prob( \hat \mu(x) \neq \mu(x))
    \end{aligned} \end{equation}
    where we used the fact that $\| \phi_{\nu^{(k)}} \|_\mathcal{H}^2 = \sigma^2$ for all $k$, which implies
    \begin{equation} \begin{aligned}
        \E_{x \sim \nu} \left[ \| \phi(x) - \mu(x) \|^2_\mathcal{H} \right] = 1 - \sigma^2
    \end{aligned} \end{equation}
    by virtue of $\kappa(x,x)=1$. Thus, we see that it is sufficient to bound the error rate of the tree. Fix a node $t$ with remaining components $N(t)$. The algorithm chooses $\bar x, i^*, k^*, l^* \in N(t)$ that maximize
    \begin{equation} \begin{aligned}
        \xi(\bar x, i,k,l) := \E_{x' \sim_{i.i.d.} \nu^{(k)}, y \sim \nu^{(l)}}[ \kappa_i(\bar x,x') - \kappa_i(\bar x,y)].
    \end{aligned} \end{equation}
    Note that $\E_{\bar x \sim \nu^{(k^*)}}[\xi(\bar x, i^*,k^*,l^*)] \ge \tau$ by definition of the axis-aligned kernel similarity. Therefore, $\xi(\bar x, i,k,l) \ge \tau$. Without loss of generality, we may assume that $N(t) = [K']$ for some $K' \le K$, and that $k^*=1, l^*=2$. For any $m \in [K']$ define
    \begin{equation} \begin{aligned}
        \zeta(m) = \E_{y \sim \nu^{(m)}}[ \kappa_{i^*}(\bar x,y)].
    \end{aligned} \end{equation}
    We may assume that $\zeta(m) \le \zeta(m+1)$ for all $2 \le m \le K'$. If this does not hold, simply sort and relabel. We know that $\zeta(2) \le \zeta(1) - \tau$. Therefore, the scalars $\{\zeta(m)\}_{m=1}^{K'}$ are spread out on an interval of length at least $\Delta \ge \tau$.
    Recall that the threshold $\theta$ is chosen exactly halfway where the jump between $\zeta(m)$ and $\zeta(m+1)$ is largest. For any mixture component $m \in N(t)$ and any $x \sim \nu^{(m)}$, the threshold cut at $\theta$ can only make a mistake (i.e. assign $x$ to the child node that erroneously does not contain index $m$) if either $\kappa_{i^*}(\bar x,x) < \theta < \zeta(m)$ or vice versa. This can only happen if
    \begin{equation} \begin{aligned}
    |\kappa_{i^*}(\bar x, x) - \zeta(m)|
    > |\theta - \zeta(m)|.
    \end{aligned} \end{equation}
    We can now use Chebyshev's inequality to bound the probability of making mistakes.
    \begin{equation} \begin{aligned}
        \Prob_{x \sim \nu} \left( x \text{ goes to wrong child at $t$}\right) &\le
        \sum_{m \in N(t)} p^{(m)} \cdot \Prob_{x \sim \nu^{(m)}}(|\kappa_{i^*}(\bar x,x) - \zeta(m)|
    > |\theta - \zeta(m)|) \\ &\le
        \sum_{m \in N(t)} p^{(m)} \cdot \frac{\epsilon^2}{|\theta - \zeta(m)|^2} \\
        &\le \frac{\alpha \epsilon^2}{K} \sum_{m \in N(t)} \frac{1}{|\theta - \zeta(m)|^2}.
    \end{aligned} \end{equation}
    Now we argue just as we did in the proof of Theorem \ref{theo:price1} (see Claim 1 from Appendix \ref{app:price1}), except that we have quadratic concentration and not exponential concentration. This is not a problem for Theorem 1.3 in \citep{farkas2025fenton}, as the (singular) kernel $K(u) = -u^{-2}$ is still concave and monotone on $(-1,0)$ and $(0,1)$. The corresponding field function is $J(s) = -s^{-2} - (1-s)^{-2}$, which is bounded from above on $[0,1]$ and finite everywhere except at the interval endpoints. Thus, an upper bound occurs when all $\zeta(m)$ are equidistant on the interval of length $\Delta$ and $N(t) = [K]$. In that case, $|\zeta(m+1)-\zeta(m)| = \frac{\Delta}{K-1}$ for all $m \le K-1$. Then, 
    \begin{equation} \begin{aligned}
        \sum_{m \in N(t)} \frac{1}{|\theta - \zeta(m)|^2} &\le 2 \sum_{k=1}^{\lceil K/2 \rceil} \frac{1}{(2k-1)^2 (\frac{\Delta}{2(K-1)})^2} \\
        &\le \frac{8 (K-1)^2}{\Delta^2} \sum_{k=1}^\infty \frac{1}{(2k-1)^2} \\
        &\le \frac{8(K-1)^2 (0.5 + \pi^2/12)}{\Delta^2}.
    \end{aligned} \end{equation}
    This implies (at every node)
    \begin{equation} \begin{aligned}
        \Prob_{x \sim \nu} \left( x \text{ goes to wrong child}\right) \le \frac{(4 + \frac{2\pi^2}{3})\alpha \epsilon^2 (K-1)^2}{K\tau^2}.
    \end{aligned} \end{equation}
    Since there are no more than $K$ nodes, we obtain
    \begin{equation} \begin{aligned}
        \Prob_{x \sim \nu}(\hat \mu(x) \neq \mu(x)) \le \frac{(4 + \frac{2\pi^2}{3})\alpha \epsilon^2 (K-1)^2}{\tau^2}.
    \end{aligned} \end{equation}
    Obviously, the probability can never exceed $1$, so we take the maximum. Plugging this back into the price of explainability yields the desired result.
\end{proof}
\end{document}